\documentclass[10pt,twocolumn,letterpaper]{article}

\usepackage[pagenumbers]{iccv} 

\usepackage{booktabs}
\usepackage{multirow}
\usepackage{diagbox}
\usepackage{pgfplots}

\definecolor{iccvblue}{rgb}{0.21,0.49,0.74}
\usepackage[pagebackref,breaklinks,colorlinks,allcolors=iccvblue]{hyperref}
\usepackage[capitalize]{cleveref}
\usepackage{amsmath}
\usepackage{algorithm}
\usepackage{algpseudocode}

\newcommand\mypara[1]{\vspace{1.0mm}\noindent\textbf{#1}}
\usepackage{booktabs}
\usepackage{multirow}
\usepackage{diagbox}
\usepackage{pgfplots}

\def\paperID{1655} 
\def\confName{ICCV}
\def\confYear{2025}

\newcommand{\ourmethod}{{\textbf{\textsc {E2ED$^2$}}}\xspace}

\definecolor{darkblue}{RGB}{0, 0, 139}

\title{\ourmethod: Direct Mapping from Noise to Data for Enhanced Diffusion Models}

\author{
Zhiyu Tan$^{1,2}$\thanks{Equal contribution.}, WenXu Qian$^{1,2}$\footnotemark[1], Hesen Chen$^{2}$, Mengping Yang, Lei Chen$^{1,2}$, Hao Li$^{1,2}$\thanks{Corresponding author.} \\
$^{1}$ {Fudan University} \quad $^{2}$ {Shanghai Academy of Artificial Intelligence for Science}
}

\begin{document}
\maketitle
\begin{abstract}
Diffusion models have established themselves as the de facto primary paradigm in visual generative modeling, revolutionizing the field through remarkable success across various diverse applications ranging from high-quality image synthesis to temporal aware video generation.
Despite these advancements, three fundamental limitations persist, including 1) discrepancy between training and inference processes, 2) progressive information leakage throughout the noise corruption procedures, and 3) inherent constraints preventing effective integration of modern optimization criteria like perceptual and adversarial loss.
To mitigate these critical challenges, we in this paper present a novel end-to-end learning paradigm that establishes direct optimization from the final generated samples to initial noises.
Our proposed End-to-End Differentiable Diffusion, dubbed \textit{\ourmethod}, introduces several key improvements: it eliminates the sequential training-sampling mismatch and intermediate information leakage
via conceptualizing training as a direct transformation from isotropic Gaussian noise to the target data distribution. 
Additionally, such training framework enables seamless incorporation of adversarial and perceptual losses into the core optimization objective.
Comprehensive evaluation across standard benchmarks including COCO30K and HW30K reveals that our method achieves substantial performance gains in terms of Fréchet Inception Distance (FID) and CLIP score, even with fewer sampling steps (less than 4).
Our findings highlight that the end-to-end mechanism might pave the way for more robust and efficient solutions, \emph{i.e.,} combining diffusion stability with GAN-like discriminative optimization in an end-to-end manner.
%
%
%
%
%
%
%
\end{abstract}

\section{Introduction}
\label{sec:intro}





Diffusion models (DMs)~\cite{ho2020denoising} have driven significant advances in visual generative modeling through its iterative denoising mechanisms, surpassing previous milestone approaches including Variational Autoencoders (VAEs)\cite{kingma2013auto} and Generative Adversarial Networks(GANs)~\cite{goodfellow2020generative}.
For instance, DMs have enabled thrilling experiences in producing high-fidelity images and animation videos conditioned on various conditions including text~\cite{peebles2023scalable,saharia2022photorealistic,ramesh2022hierarchical,betker2023improving,esser2403scaling,yang2024cogvideox,ma2024latte,tan2024empirical}, image frames~\cite{singer2022make,ho2022video,sun2024hunyuan,gupta2024photorealistic,hu2024animate}, camera trajectory~\cite{poole2022dreamfusion,hoppe2022diffusion,zheng2024cami2v,he2024cameractrl}, \emph{etc.}
However, one of the key challenges of employing DMs in practical applications is their computational iterative denoising process~\cite{song2023consistency, lu2022dpm, karras2024analyzing}, making it slow and expensive.
Recently, consistency models (CMs)~\cite{song2023consistency} have attracted extensive attention for their capability of aligning intermediate representations throughout the denoising process, enabling faster inference with enhanced stability and reliability.
These developments reveal persistent trade-offs between output quality, inference efficiency, and generation consistency.

Despite these tremendous achievements, DMs still face several fundamental limitations in their theoretical framework and practical implementation.
\textbf{1) Training-Inference Discrepancy}: the inherent divergence between single-step denoising optimization (training) and multi-step iterative generation (inference) creates a critical performance gap.
Conventionally, DMs is trained to optimize individual step predictions with the evidence lower bound (ELBO) objective:
$\mathcal{L}_{\text{ELBO}} = \mathbb{E}_{t,\epsilon}\left[\|\epsilon - \epsilon_\theta(x_t,t)\|^2\right]$,
without considering temporal error propagation in the reverse process. 
This leads to error accumulation that scales with $O(\sqrt{T})$ for $T$ sampling steps~\cite{ho2020denoising, song2023improved}, particularly detrimental when using accelerated sampling schedules with $T < 50$.
\textbf{2) Information Leakage of Training and Sampling Noise Schedule:}
the theoretical validity of diffusion processes relies on the terminal state $x_T$ converging to isotropic Gaussian noise ($\mathcal{N}(0,I)$) through the forward process: $q(x_T|x_0) = \prod_{t=1}^T q(x_t|x_{t-1})$.
However, practical implementations with finite $T$ steps exhibit non-zero mutual information $I(x_T;x_0) > 0$, making $x_T$ deviates from the true Gaussian noise, leading to information leakage, quantified by the KL divergence $D_{\text{KL}}(q(x_T) \| \mathcal{N}(0,I)) > 0$. Such deviation compromises the model’s ability to accurately reconstruct data, introducing a biased initialization space that degrade the quality of the generated outputs.
\textbf{3) Incompatibility with Other Optimization Objectives:}
the standard training paradigm's stochastic time step sampling ($\{t\} \sim \mathcal{U}[1,T]$) fundamentally conflicts with trajectory-sensitive loss functions, such as the perceptual loss~\cite{liu2021generic} and the adversarial GAN loss~\cite{goodfellow2020generative}.
These optimization losses are crucial for improving the perceptual quality and semantic coherence of synthesized images. For instance, incorporating the adversarial loss could effectively enhance the visual details and rich appearance~\cite{ledig2017photo,wang2018esrgan}.
However, the inherent step-wise optimization of diffusion models prevents integration of these loss enhancements, thereby restricting their potential to achieve superior visual fidelity and high-quality synthesis, particularly for tasks requiring fine-grained detail generation.
These limitations hinder the performance, consistency, and flexibility of diffusion models, highlighting the necessity for innovations that address these challenges while maintaining their generative strengths.

In order to overcome the aforementioned limitations, we introduce an \textbf{E}nd-to-\textbf{E}nd \textbf{D}ifferentiable \textbf{D}iffusion (\textbf{\ourmethod}) modelthat aligns the training and sampling processes while addressing fundamental challenges.
Unlike conventional methods that primarily aim to predict noise at randomly sampled intermediate states, our approach directly optimizes the ultimate reconstruction task. Specifically, it transforms pure Gaussian noise into the target distribution through a unified process, thereby bridging the gap between training and sampling. 
This framework enables the model to effectively learn how to manage cumulative errors across all sampling steps. Furthermore, it resolves information leakage in $x_T$ by treating the training process as a direct mapping from pure noise to the data distribution, ensuring a more robust and principled learning paradigm.
Additionally, the unified objective of our framework also facilitates the incorporation of perceptual and adversarial losses, thereby enhancing the fidelity and semantic coherence of the generated outputs. 
Importantly, these improvements are achieved without compromising theoretical soundness or stability, ensuring consistency and robustness throughout the generative process.

We validate the effectiveness of our proposed approach through extensive experiments conducted on well-established benchmarks, including COCO30K \cite{lin2014microsoft} and HW30K \cite{chen2024pixart_delta}. 
Our method consistently outperforms existing state-of-the-art diffusion models across key evaluation metrics such as Fréchet Inception Distance (FID) \cite{heusel2017gans_fid} and CLIP score \cite{radford2021clip}.
Notably, our framework demonstrates superior performance while requiring fewer sampling steps (less than $4$), illustrating its efficiency and scalability. 
These results highlight the transformative potential of end-to-end training frameworks in advancing diffusion-based generative models.
By explicitly addressing the core shortcomings of existing methods, our approach sets a new benchmark for efficiency, quality, and consistency in generative modeling.
To sum up, our primary contributions are three-fold: 
\begin{itemize}
    \item We propose \textbf{\ourmethod}, a unified end-to-end training framework for diffusion models. By directly transforming pure noise into the target data distribution, our method mitigates the training-sampling gap and provides a generalizable framework.
    \item Our proposed method is conceptually simple and orthogonal to various optimization objectives, allowing seamless incorporation of perceptual and adversarial losses for enhanced synthesis quality.
    \item Extensive experiments demonstrate the effectiveness of our proposed method. Notably, our method achieves strong performance with fewer sampling steps, improving efficiency without compromising generation quality.
\end{itemize}

\section{Related Work}
\subsection{Generative models.} Generative models have undergone significant evolution, starting from Variational Autoencoders (VAEs)\cite{kingma2013auto} and Generative Adversarial Networks (GANs)\cite{goodfellow2020generative}. VAEs leverage probabilistic latent variable models to generate structured data but often struggle with producing high-fidelity samples due to their reliance on explicit likelihood optimization. GANs, on the other hand, use adversarial training to synthesize visually appealing data but are prone to instability and mode collapse, limiting their scalability. Modern advancements, such as diffusion models and latent generative approaches, have addressed some of these issues by introducing iterative refinement and efficient latent representations. However, these methods often involve complex training dynamics and lack alignment between training and inference phases\cite{ho2020denoising,songdenoising,rombach2022high}. Unlike these methods, our end-to-end approach directly optimizes the generative trajectory, ensuring consistency between training and sampling while reducing complexity.

\subsection{Diffusion Models} 
Diffusion models, such as Denoising Diffusion Probabilistic Models (DDPM)~\cite{ho2020denoising} and their extensions like stochastic differential equations (SDE)~\cite{song2020score} and ordinary differential equations (ODE)~\cite{zhou2024fast, lu2022dpm}, have set the benchmark for generative modeling across various tasks. These models operate by progressively transforming Gaussian noise into the target data distribution through a multi-step denoising process, leveraging a learned sequence of reverse transformations. Despite their remarkable success, diffusion models face a significant limitation stemming from a fundamental training-sampling mismatch. During training, the model is optimized to predict noise in a single-step denoising task for randomly sampled time steps. However, in the sampling phase, the model must iteratively refine the data across multiple steps~\cite{songdenoising,ho2020denoising,karras2024analyzing}. This discrepancy introduces a training-sampling gap, leading to compounded errors and inefficiencies during inference.
Our proposed method addresses these limitations by unifying the training and sampling processes through an end-to-end optimization framework. Instead of focusing on intermediate noise predictions, we directly optimize the final reconstruction, aligning the model's training objectives with the sampling procedure. This not only bridges the training-sampling gap but also mitigates error accumulation and enhances inference efficiency. By directly targeting the quality of the generated outputs, our method achieves superior performance, paving the way for more robust diffusion-based generative models.


\subsection{Accelerating DMs.}
Efforts to accelerate diffusion models have introduced several innovative techniques. Latent diffusion models (LDM)\cite{rombach2022high} reduce computation by operating in a compressed latent space, while progressive distillation minimizes the number of sampling steps through staged knowledge transfer\cite{salimans2022progressive, lin2024sdxl}. Consistency models align intermediate representations to facilitate fast sampling~\cite{song2023consistency, song2023improved}, and latent consistency models extend this concept to latent spaces for further efficiency gains~\cite{luo2023latent}. Several other acceleration techniques have also been proposed, such as knowledge distillation from pre-trained diffusion models~\cite{meng2023distillation} and analytical estimations of sampling trajectories~\cite{bao2022analytic}. However, these approaches often involve additional assumptions, such as specific network architectures~\cite{karras2022elucidating} or auxiliary training losses~\cite{nichol2021improved}, and may still rely on multi-step sampling procedures~\cite{xiao2021tackling}. Our method simplifies the generative process by learning the entire trajectory in an end-to-end manner, eliminating reliance on step-wise refinements and achieving robust performance with fewer constraints.

\begin{figure*}[th]
    \centering
    \includegraphics[width=\textwidth]{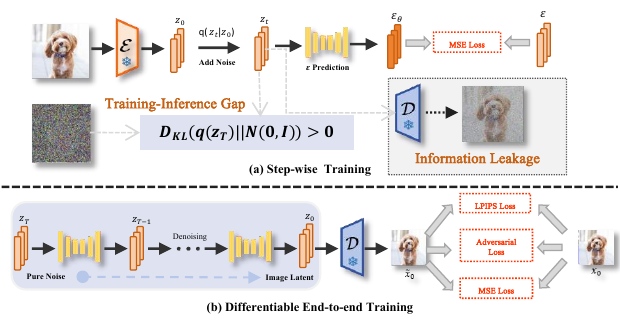} 
    \caption{
        \textbf{Comparison of two training methods for diffusion models.} (a) illustrates the \textbf{single-step training method}, where the model is trained to predict noise in a single denoising step for randomly sampled time steps. This approach introduces a training-sampling gap, as the training focuses on single-step denoising, whereas testing requires iterative multi-step denoising. Furthermore, the forward noising process can result in information leakage, where the final state \(x_T\) deviates from ideal Gaussian noise, compromising the reconstruction quality. (b) illustrates our proposed \textbf{end-to-end training method}, \ourmethod, which directly optimizes the entire sampling trajectory. Beginning with pure Gaussian noise, the model generates images through multi-step sampling, aligning the training and testing processes seamlessly. This approach effectively eliminates information leakage and enables the integration of advanced loss functions, such as perceptual and GAN losses, thereby improving the fidelity, semantic consistency, and overall quality of the generated images.
    }
    \label{fig:end_to_end_trajectory}
\end{figure*}

\begin{algorithm}
\caption{End-to-End Training in Latent Space for Text-to-Image Diffusion with EMA}
\begin{algorithmic}[1]
\Require 
$\theta$: Model parameters; 
$\theta_{\text{EMA}}$: EMA model parameters; 
$\tau$: EMA decay rate; 
$T$: Number of diffusion steps; 
$d(\cdot, \cdot)$: Reconstruction loss metric; 
$D$: Paired image-caption training dataset $\{(\mathbf{x}_0, \mathbf{c})\}$; 
$\alpha_t, \sigma_t$: Noise schedule parameters; 
$\theta_{\text{pre}}$: Pre-trained model parameters.
\Ensure Optimized model parameters $\theta^*$ and EMA parameters $\theta^*_{\text{EMA}}$.
\State Initialize: $\theta \gets \theta_{\text{pre}}$ and $\theta_{\text{EMA}} \gets \theta$.
\While{\textit{not converged}}
    \State Sample $(\mathbf{x}_0, \mathbf{c}) \sim D$;
    \State Encode $\mathbf{x}_0$ to latent space: $\mathbf{z}_0 = E(\mathbf{x}_0)$;
    \State Encode caption $\mathbf{c}$: $\mathbf{e}_\mathbf{c} = T(\mathbf{c})$;
    \State Sample initial noise: $\mathbf{z}_T \sim \mathcal{N}(\mathbf{0}, \mathbf{I})$;
    \For{$t = T \text{ to } 1$}
        \State Sample $\mathbf{z} \sim \mathcal{N}(\mathbf{0}, \mathbf{I})$;
        \State Compute $\hat{\mathbf{z}}_t \gets \mathbf{z}_t + \sqrt{\alpha_t^2 - \sigma_t^2} \cdot \mathbf{z}$;
        \State Predict the denoised latent: $\mathbf{z}_t \gets G_\theta(\hat{\mathbf{z}}_t, \mathbf{e}_\mathbf{c}, t)$;
    \EndFor
    \State Set $\hat{\mathbf{z}}_0 = \mathbf{z}_t=0$ (final predicted latent variable);
    \State Compute reconstruction loss: $L_{\text{recon}} = d(\mathbf{z}_0, \hat{\mathbf{z}}_0)$;
    \State Backpropagate and update $\theta$: $\theta \gets \theta - \eta \nabla_\theta L_{\text{recon}}$;
    \State Update EMA parameters: $\theta_{\text{EMA}} \gets \text{stopgrad}(\tau \theta_{\text{EMA}} + (1 - \tau) \theta)$;
\EndWhile
\Ensure Optimized model parameters $\theta^*$ and EMA parameters $\theta^*_{\text{EMA}}$
\end{algorithmic}
\end{algorithm}

\section{Method}
\label{sec:method}

\subsection{Preliminary}

\noindent \textbf{Diffusion Models.}
The Diffusion Model\cite{ho2020denoising} serves as the foundational framework. It defines a forward process that progressively adds Gaussian noise to data samples $\mathbf{x}_0 \sim q(\mathbf{x}_0)$ over $T$ time steps, modeled as:
\begin{equation}
q(\mathbf{x}_t | \mathbf{x}_{t-1}) = \mathcal{N}(\mathbf{x}_t; \sqrt{1 - \beta_t} \mathbf{x}_{t-1}, \beta_t \mathbf{I}),
\end{equation}
where $\beta_t \in (0, 1)$ are noise variance schedules. The reverse process, which aims to reconstruct clean data, is parameterized as:
\begin{equation}
p_\theta(\mathbf{x}_{t-1} | \mathbf{x}_t) = \mathcal{N}(\mathbf{x}_{t-1}; \mu_\theta(\mathbf{x}_t, t), \Sigma_\theta(\mathbf{x}_t, t)).
\end{equation}
During training, the model learns to predict the noise $\epsilon$ added to the data $\mathbf{x}_t$ at each time step by optimizing the objective:
\begin{equation}
L(\theta) = \mathbb{E}_{q(\mathbf{x}_0, \mathbf{x}_t, \epsilon)} \left[ \|\epsilon - \epsilon_\theta(\mathbf{x}_t, t)\|^2 \right].
\end{equation}
Crucially, this loss function optimizes the model for a single denoising step, assuming independence between steps. However, during sampling, the model performs multi-step denoising, where each step depends on the accuracy of the preceding steps, creating a coupling between time steps. This mismatch between single-step training and multi-step sampling often limits the quality of the generated samples, as errors can accumulate across steps.

\noindent \textbf{Latent Diffusion Model (LDM)} \cite{rombach2022high} addresses the computational inefficiency of operating in the high-dimensional data space by introducing a latent representation $\mathbf{z}_0 = \mathcal{E}(\mathbf{x}_0)$, obtained via an encoder $\mathcal{E}$. The forward and reverse diffusion processes are then performed in the latent space:
\begin{equation}
q(\mathbf{z}_t | \mathbf{z}_{t-1}) = \mathcal{N}(\mathbf{z}_t; \sqrt{1 - \beta_t} \mathbf{z}_{t-1}, \beta_t \mathbf{I}),
\end{equation}
\begin{equation}
p_\theta(\mathbf{z}_{t-1} | \mathbf{z}_t) = \mathcal{N}(\mathbf{z}_{t-1}; \mu_\theta(\mathbf{z}_t, t), \Sigma_\theta(\mathbf{z}_t, t)).
\end{equation}
The training objective remains similar:
\begin{equation}
L(\theta) = \mathbb{E}_{q(\mathbf{z}_0, \mathbf{z}_t, \epsilon)} \left[ \|\epsilon - \epsilon_\theta(\mathbf{z}_t, t)\|^2 \right].
\end{equation}
While LDM reduces computational costs by operating in a lower-dimensional space, it inherits the same limitation as diffusion models: the training is optimized for a single denoising step, while the sampling process involves multi-step refinement, where errors can propagate and degrade the final result.

\subsection{The Proposed \ourmethod}

To address the aforementioned limitations of diffusion models, particularly the mismatch between single-step training and multi-step sampling, we propose an \textbf{end-to-end training approach} that directly optimizes the generation process from pure Gaussian noise $\mathbf{z}_T$ to the reconstructed latent representation $\mathbf{z}_0$, conditioned on textual descriptions. This approach simplifies the training objective and enables joint optimization over the entire sampling trajectory in the latent space, effectively aligning training and sampling objectives.

In traditional diffusion models, the training objective distributes the loss across all $T$ timesteps, with the model trained to predict the noise $\epsilon$ added at each intermediate timestep $\mathbf{z}_t$. This involves $T$ separate losses during training, while the final generative performance is indirectly affected by the coupling of intermediate timesteps during sampling. In contrast, our method focuses on the ultimate reconstruction quality of $\mathbf{z}_0$ by training the model end-to-end with a single loss. Specifically, we redefine the training objective by applying the loss directly to the reconstructed latent result $\hat{\mathbf{z}}_0$, as follows:
\begin{equation}
L_{\text{recon}}(\theta) = \mathbb{E}_{q(\mathbf{z}_T , \mathbf{z}_0)} \left[ d(\mathbf{z}_0, \hat{\mathbf{z}}_0) \right],
\end{equation}
where $\hat{\mathbf{z}}_0$ is the output of the generative process starting from pure Gaussian noise $\mathbf{z}T$ and conditioned on the text embedding. The metric function $d(\cdot, \cdot)$ quantifies the distance between the ground truth latent representation $\mathbf{z}_0$ and the reconstructed output $\hat{\mathbf{z}}_0$. By operating in the latent space instead of the pixel space, our approach leverages pre-trained image encoders for efficient and semantically meaningful representations.

The training process, outlined in Algorithm~\ref{alg:text_to_image_training_ema}, involves learning a mapping from a noisy latent input $\mathbf{z}_T$ to the clean latent representation $\mathbf{z}_0$, conditioned on a textual embedding $\mathbf{e}_\mathbf{c}$ obtained from a pre-trained text encoder. To ensure stability and improve model performance, Exponential Moving Average (EMA) is employed to update the model parameters incrementally.

A critical aspect of training lies in defining the metric function $d(\cdot, \cdot)$, which measures the similarity between the predicted latent representation $\hat{\mathbf{z}}_0$ and the target $\mathbf{z}_0$. Several choices are commonly used, depending on the specific goals of the task. The \textbf{L1 Loss}, defined as $d_{\text{L1}}(\mathbf{z}_0, \hat{\mathbf{z}}_0) = \|\mathbf{z}_0 - \hat{\mathbf{z}}_0\|_1$, calculates the mean absolute error (MAE) and is known for its robustness to outliers. In contrast, the \textbf{L2 Loss}, $d_{\text{L2}}(\mathbf{z}_0, \hat{\mathbf{z}}_0) = \|\mathbf{z}_0 - \hat{\mathbf{z}}_0\|_2^2$, computes the mean squared error (MSE), which penalizes larger deviations more strongly, potentially improving convergence but being more sensitive to outliers. Another widely used metric is the \textbf{Learned Perceptual Image Patch Similarity (LPIPS) Loss}, expressed as $d_{\text{LPIPS}}(\mathbf{z}_0, \hat{\mathbf{z}}_0) = \text{LPIPS}(\mathbf{z}_0, \hat{\mathbf{z}}_0)$. This loss leverages features from pre-trained neural networks to better align with human perceptual judgments of visual similarity, making it particularly effective in improving output quality for image-generation tasks. Additionally, a \textbf{hybrid metric} can be employed to combine these approaches, with a weighted formulation $d(\mathbf{z}_0, \hat{\mathbf{z}}_0) = \lambda_{\text{L1}} d_{\text{L1}} + \lambda_{\text{L2}} d_{\text{L2}} + \lambda_{\text{LPIPS}} d_{\text{LPIPS}}$, where weights $\lambda_{\text{L1}}, \lambda_{\text{L2}}, \lambda_{\text{LPIPS}}$ balance their contributions, enabling the model to leverage the complementary strengths of these loss functions.

The training process samples a random timestep \( t \sim \{1, \dots, T\} \) per iteration to construct noisy samples \( \mathbf{z}_t \). Unlike conventional stepwise training, our approach applies the reconstruction loss solely to the final output \( \hat{\mathbf{z}}_0 \), enabling end-to-end optimization and mitigating error propagation. The flexibility of \( L_{\text{recon}} \) allows integration of various metrics \( d(\cdot, \cdot) \), such as LPIPS for perceptual quality and L1/L2 for structural accuracy. The impact of these choices is analyzed in the experimental section.

\noindent \textbf{Combination with adversarial GAN losses.}
One another advantage of this approach is its flexibility to incorporate auxiliary objectives, such as adversarial training. By introducing a GAN-based loss $L_{\text{GAN}}$, the model can further improve the perceptual quality of the generated data. The combined training objective becomes:
\begin{equation}
L_{\text{total}}(\theta) = L_{\text{recon}}(\theta) + \lambda_{\text{GAN}} L_{\text{GAN}}(\theta),
\end{equation}
where $\lambda_{\text{GAN}}$ controls the weight of the adversarial loss. The reconstruction loss ensures faithful reproduction of the data, while the adversarial loss encourages high-quality and realistic outputs. This approach addresses the core limitations of existing diffusion models, providing a unified and flexible framework for efficient and high-quality generation.
\section{Experiment}
\label{sec:experiment}

\subsection{Experimental Setting}

\subsubsection{Datasets}
To enhance the aesthetic quality and semantic consistency of the dataset, we began by integrating the COYO dataset\cite{carlini2024poisoning_coyo} with an internally curated collection and conducting rigorous screening. Only images with aesthetic scores exceeding 5.5 were retained, ensuring superior visual quality. To generate captions, we employed the LLaVA 13B model\cite{liu2024visual}, which leverages robust language understanding and image description generation capabilities to produce highly accurate and semantically consistent captions automatically. After filtering and captioning, we selected 120k high-quality images to form our final training dataset.

\subsubsection{Baselines}
To assess our method’s performance, we compare it with state-of-the-art text-to-image models, including SDXL~\cite{sdxl}, SDXL-Turbo, SDXL-Lightning~\cite{lin2024sdxl}, Latent Consistency Model (LCM)~\cite{luo2023latent}, and PixArt-\(\delta\)~\cite{chen2024pixart_delta}. SDXL excels in high-resolution image generation, while SDXL-Turbo and SDXL-Lightning optimize inference speed with minimal quality loss. PixArt-\(\delta\) and LCM reduce sampling steps via Latent Consistency Modeling, offering efficient generation with competitive visual fidelity. These comparisons highlight trade-offs between efficiency, scalability, and quality, where our method demonstrates significant advantages.

\subsubsection{Evaluation Metrics} 
To validate the effectiveness of our approach, we employed three metrics to assess the quality of generated images and their alignment with textual descriptions: FID on COCO30k, FID on HW30k, and CLIP score. FID evaluates the distributional similarity between real and generated data, with separate assessments on the COCO30k and HW30k datasets. The CLIP score measures the semantic correspondence between images and their associated captions. For fair comparisons, we followed the evaluation methodology introduced in PixArt-\(\delta\)~\cite{chen2024pixart_delta}. Specifically, FID evaluations were conducted on HW30k, a dataset introduced in PixArt-\(\delta\), containing 30k high-quality text-image pairs with remarkable aesthetic value and precise semantic annotations. Additional evaluations on the widely used COCO30k dataset provided consistent benchmarking against existing generative models. These metrics collectively highlight the superior perceptual quality and semantic alignment achieved by our approach.

\subsubsection{Implementation Details}
Our model was initialized using the pretrained PixArt-\(\delta\) model\cite{chen2024pixart_delta} to leverage its robust text-to-image generation capabilities. The training process was carried out on 64 NVIDIA A100 GPUs, utilizing a batch size of 8 and a total of 24k training steps. We adopted the AdamW optimizer, configured with a weight decay of \(1 \times 10^{-8}\) and a fixed learning rate of \(1 \times 10^{-6}\), ensuring effective gradient updates while mitigating overfitting. To maintain stable and smooth updates of model weights, an Exponential Moving Average (EMA) with a decay coefficient of 0.95 was employed throughout the training process. The experiments were conducted on an internal dataset comprising 120k high-quality images, specifically curated to cover a diverse range of subjects and styles, providing a robust foundation for training. This setup enabled our model to achieve consistent performance improvements across all evaluation metrics.

\subsection{Main Results}

\begin{table}[t]
    \centering
    \caption{
    \textbf{Quantitative comparison of our method with state-of-the-art approaches on text-to-image synthesis tasks at $1024 \times 1024$ resolution}. This table compares various methods based on COCO FID, HW FID, and CLIP score, showcasing the effectiveness of our approach in generating high-quality and semantically aligned images.
    }
    \label{tab:sota_comp}
    \resizebox{\columnwidth}{!}{%
    \begin{tabular}{lccccc}
    \toprule
    \multirow{2}{*}{Method} & \multirow{2}{*}{Model Size} & \multirow{2}{*}{NFE} & \textbf{COCO} & \textbf{HW} & \textbf{COCO} \\
    & & & \textbf{FID} $\downarrow$ & \textbf{FID} $\downarrow$ & \textbf{CLIP} $\uparrow$ \\
    \midrule
    SDXL~\cite{sdxl}                  & 2.5B  & 32 & 14.28 & 7.96  & 31.68 \\
    SDXL-Turbo                        & 2.5B  & 4  & 20.66 & 14.98 & 31.65 \\
    SDXL-Lightning~\cite{lin2024sdxl} & 2.5B  & 4  & 21.60 & 11.20 & 31.26 \\
    LCM~\cite{luo2023latent}          & 0.86B & 4  & 31.31 & 34.91 & 30.30 \\
    PixArt-\(\delta\)~\cite{chen2024pixart_delta} & 0.6B & 4 & 28.49 & 11.30 & 30.83 \\
    \ourmethod                        & 0.6B  & 4  & 25.27 & 9.76  & 32.76 \\
    \midrule
    PixArt-\(\delta\)~\cite{chen2024pixart_delta} & 0.6B & 3 & 28.29 & 11.07 & 30.80 \\ 
    \ourmethod                        & 0.6B  & 3  &  26.35 & 9.97  & 30.75 \\
    \bottomrule
    \end{tabular}%
    }
\end{table}

\subsubsection{Quantitative Results}
To evaluate the effectiveness of our proposed method and compare it against state-of-the-art approaches, we conducted a comprehensive quantitative experiment. The quantitative comparison in Table~\ref{tab:sota_comp} highlights the significant advantages of our method over state-of-the-art approaches across multiple evaluation metrics, demonstrating the effectiveness of our end-to-end training strategy. Unlike the compared methods, which rely on distillation techniques, our approach employs a small model trained entirely end-to-end, aligning the training process more closely with the sampling process and effectively bridging the gap between training and inference.
In terms of image quality, as measured by FID, our method achieves superior performance on both COCO and HW datasets. The COCO FID underscores our model’s ability to generate realistic images with fewer visual artifacts, while the HW FID demonstrates its capacity to produce aesthetically rich outputs. For text-image alignment, evaluated by CLIP score, our method achieves the highest score, reflecting stronger semantic consistency between the generated images and input prompts. This improvement can be attributed to the integration of LPIPS and GAN losses, which enhance both perceptual quality and semantic alignment.
Moreover, despite having a significantly smaller model size and requiring only four sampling steps, our method surpasses larger models like SDXL in both image quality and alignment. These results validate the strength of our framework, which successfully balances high-quality generation, perceptual detail, and semantic accuracy, establishing itself as an efficient and scalable solution for text-to-image synthesis tasks.

\subsubsection{Qualitative Results}

\begin{figure*}[htbp]
    \centering
    \includegraphics[width=\textwidth]{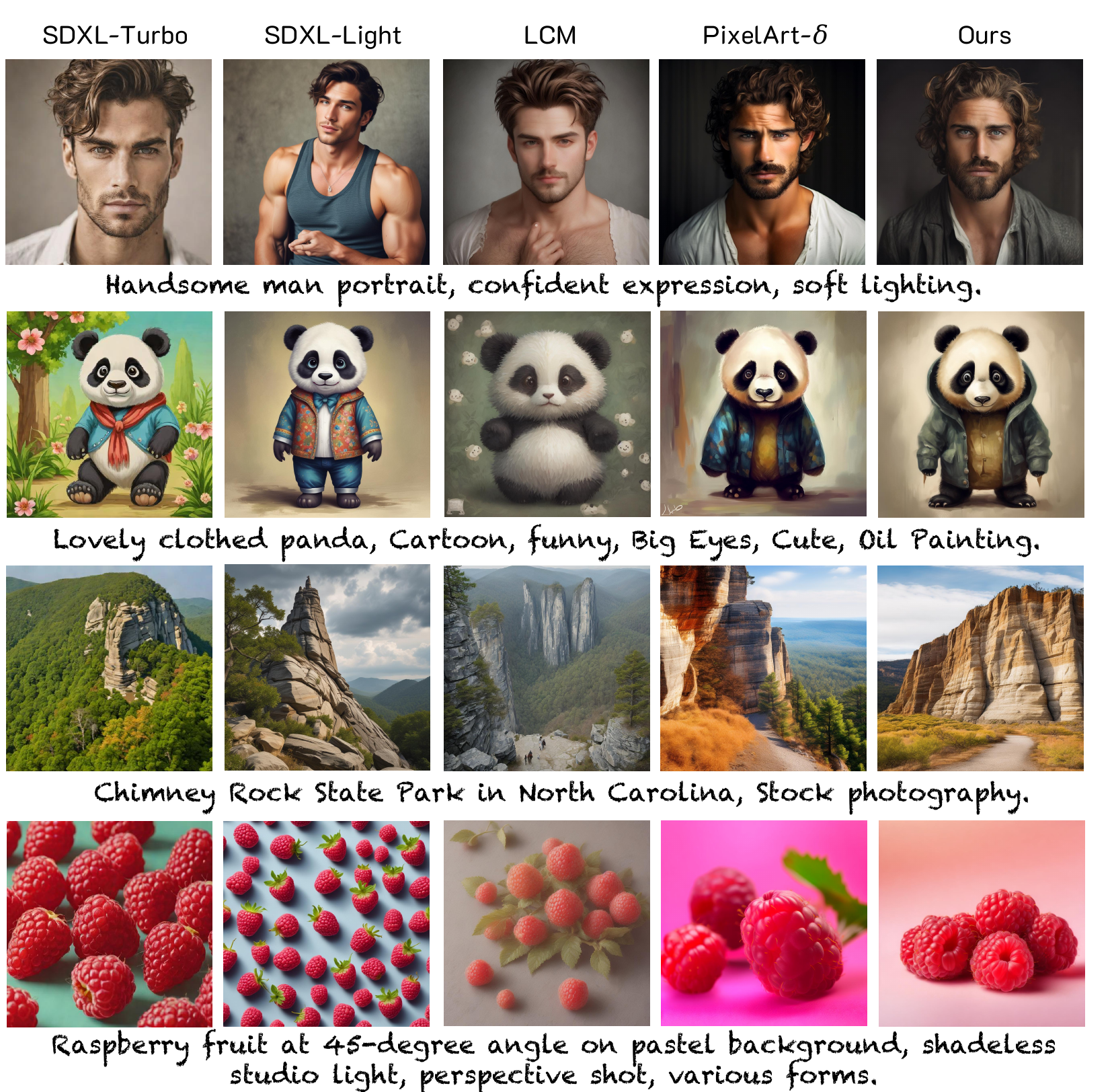} 
    \caption{
    \textbf{Qualitative comparison of synthesized images across different methods on various subjects (human portraits, objects) and styles}. Our method demonstrates superior performance in terms of image quality, aesthetic appeal, and text-image alignment. The generated images show finer details, richer textures, and better adherence to the input prompts compared to SOTA methods. This highlights the effectiveness of our end-to-end training framework and loss function design in balancing perceptual quality and semantic consistency.
    }
    \label{fig:visual_comparison_sota}
\end{figure*}

To evaluate the effectiveness of our method in generating high-quality, aesthetically appealing, and semantically aligned images, we conducted a qualitative comparison with state-of-the-art approaches. Figure~\ref{fig:visual_comparison_sota} presents the results across a variety of subjects, including human portraits and objects, in diverse styles. Our method demonstrates clear advantages in image quality, aesthetic appeal, and text-image alignment. The generated outputs exhibit finer details and sharper textures, such as individual hair strands, facial expressions, and material properties, resulting in more realistic and visually compelling images. Additionally, our approach consistently produces aesthetically superior outputs, particularly in stylized image generation tasks, where it effectively balances artistic style and content without sacrificing compositional harmony. In terms of semantic consistency, our method achieves stronger alignment between the generated images and input text prompts, benefiting from the integration of LPIPS and GAN losses, which enhance both perceptual fidelity and semantic relevance. These qualitative results, alongside quantitative findings, validate the effectiveness of our end-to-end training framework and loss function design, showcasing a robust balance between detail richness, aesthetic quality, and text-image alignment. This establishes our method as a versatile and efficient solution for diverse text-to-image synthesis tasks.

\subsubsection{Ablation Study}

\begin{table}[t]
\centering
\caption{ \textbf{Ablation study of different loss combinations.} 
This experiment investigates the contribution of various loss components (L1, L2, LPIPS, and GAN) to the overall model performance. By selectively enabling or disabling these components, we analyze their impact on the image quality and text-image alignment of generated images. The results highlight how different loss combinations influence the final output, providing insights into the effectiveness of each component.}
\resizebox{\columnwidth}{!}{%
\begin{tabular}{cccc|ccc}
\toprule
\textbf{L1} & \textbf{L2} & \textbf{LPIPS} & \textbf{GAN} & \textbf{COCO FID} $\downarrow$ & \textbf{HW FID} $\downarrow$ & \textbf{COCO CLIP} $\uparrow$ \\
\midrule
\checkmark &            &            &            &  25.61 & 10.15 & 31.32 \\
           & \checkmark &            &            &  25.34 & 9.91 & 31.31 \\
           &            & \checkmark &            &  26.68 & 10.11 & 31.24 \\
           & \checkmark & \checkmark &            &  25.27 & 9.76 & 32.76 \\
           & \checkmark & \checkmark & \checkmark &  25.74 & 9.92 & 31.75 \\
\bottomrule
\end{tabular}%
}
\label{tab:ablation_study}
\end{table}

To investigate the impact of different loss functions employed in our model, we specifically aim to understand the effects of incorporating perceptual loss (LPIPS) and GAN loss into the training process alongside baseline reconstruction losses such as L1 and L2. We hypothesize that combining multiple loss functions within our end-to-end training framework can lead to improved image quality, enhanced text-image alignment, and greater perceptual realism compared to relying on single loss functions.

We conduct the ablation study using five distinct loss configurations. The first configuration uses L1 loss alone as a baseline, focusing on pixel-level reconstruction accuracy. The second replaces L1 with L2 loss, which offers a slightly different penalty for pixel-level discrepancies. The third configuration incorporates perceptual loss (LPIPS) to enhance the perceptual quality of generated images by considering feature-level similarities. The fourth combines LPIPS with L2 loss to balance perceptual quality and pixel-level fidelity. Finally, the fifth configuration integrates GAN loss alongside LPIPS and L2 losses, aiming to improve texture realism and introduce fine-grained details. To evaluate the performance of these configurations, we measure COCO FID, HW FID, and COCO CLIP score on standard text-to-image benchmarks.

\textbf{Quantitative Evaluation.}
The experimental results reveal several important insights into the impact of different loss functions on model performance. L2 loss demonstrates marginal advantages over L1 loss in pixel-level reconstruction, serving as a slightly more effective baseline. Incorporating LPIPS loss alone does not significantly improve performance but becomes highly effective when combined with L2 loss, resulting in enhanced perceptual quality and alignment across all metrics. Adding GAN loss to the combination of LPIPS and L2 introduces high-frequency details and textures, enriching the visual realism of generated images. However, this improvement comes with a slight trade-off in overall generation quality and text-image alignment.
These observations highlight three key conclusions. First, our end-to-end training framework seamlessly integrates LPIPS and GAN losses, significantly improving image quality and text-image alignment. Second, while GAN loss enriches visual detail and texture, it introduces a trade-off between perceptual detail and metric performance. Lastly, combining multiple loss functions consistently outperforms single-loss configurations, demonstrating the value of leveraging complementary loss objectives. Together, these findings underscore the flexibility and effectiveness of our framework, providing a strong foundation for future advancements in generative modeling.

\begin{figure}[th]
    \centering
    \includegraphics[width=1.0\columnwidth]{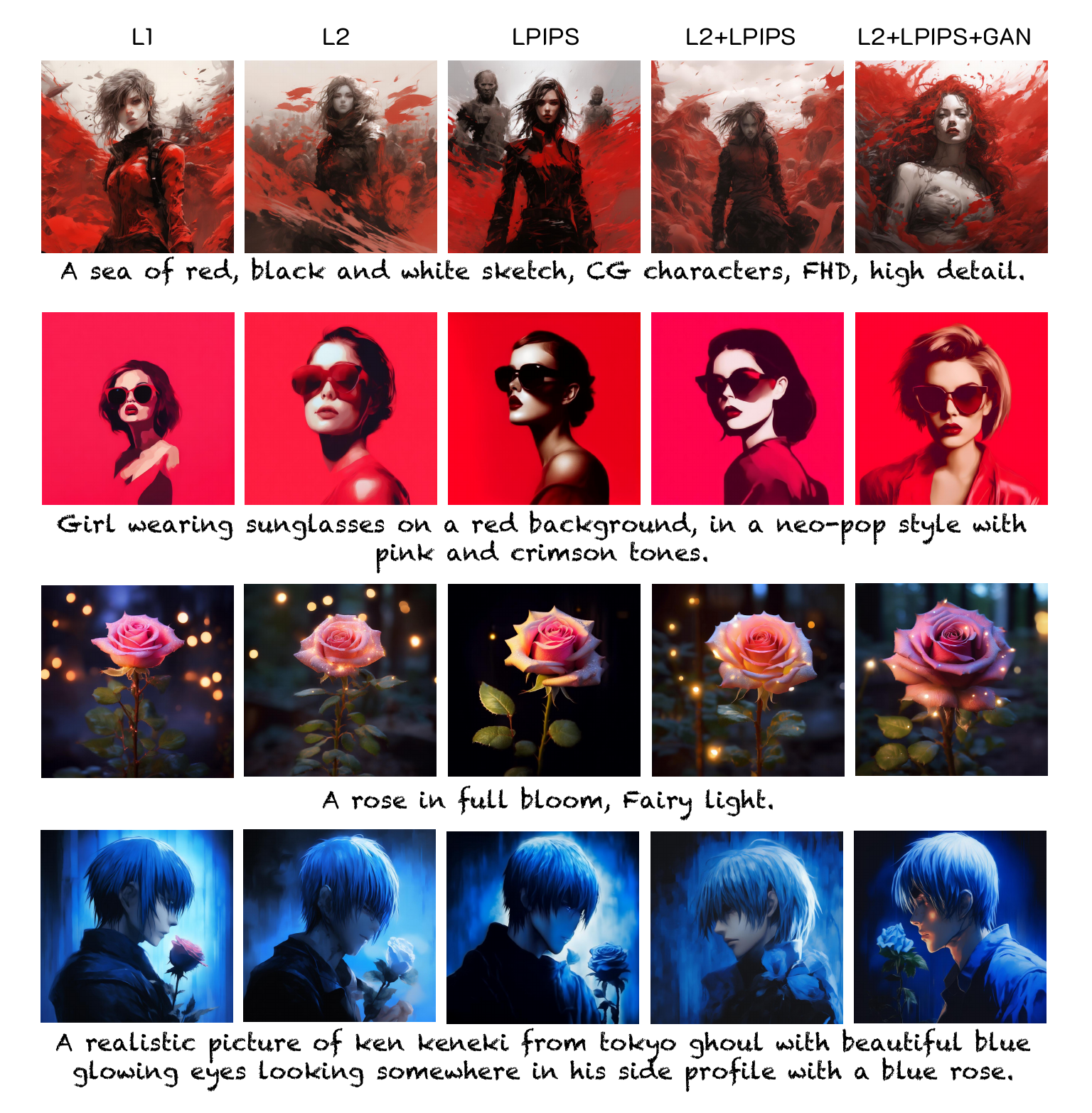} 
    \caption{
    \textbf{Qualitative comparison of generated results across different loss configurations}. 
    Each column represents a specific loss setting: L1, L2, LPIPS, L2+LPIPS, and L2+LPIPS+GAN. The inclusion of LPIPS loss improves perceptual quality, while combining L2+LPIPS with GAN loss adds high-frequency details, such as fine hair strands in portraits and intricate petal textures in flowers. Although this introduces a slight trade-off in text-image alignment, the combination of L2, LPIPS, and GAN losses achieves the best balance, producing realistic and semantically aligned outputs.}
    \label{fig:ablation_visualization}
\end{figure}

\textbf{Qualitative Evaluation.}
To further analyze the effects of different loss functions, we present qualitative comparisons of the generated outputs for flowers and human portraits, as shown in Figure~\ref{fig:ablation_visualization}. The visualizations highlight the impact of GAN loss in generating high-frequency details. For example, in the generated human portraits, the inclusion of GAN loss results in finer details such as individual strands of hair, which significantly enhances the visual realism of the output. Similarly, in the flower images, the petal textures are noticeably richer and more detailed when GAN loss is used.
These visual observations align with the quantitative results discussed earlier. While the addition of GAN loss slightly compromises overall FID and CLIP score, it introduces valuable perceptual details, enriching the generated images’ realism. This trade-off underscores the complementary nature of combining LPIPS, L2, and GAN losses, which achieves a balance between perceptual detail and semantic alignment, as evident in both the quantitative and qualitative results. Together, these findings demonstrate the strength and flexibility of our end-to-end training framework for text-to-image generation tasks.

\section{Conclusion}
\label{sec:conclusion}
Diffusion models have established themselves as a leading framework for generative modeling, delivering state-of-the-art performance across a variety of tasks. However, their potential is constrained by key limitations, including the training-sampling gap, information leakage in the noising process, and the inability to leverage advanced loss functions during training. In this work, we address these challenges through a novel end-to-end training framework that directly optimizes the full sampling trajectory. By aligning the training and sampling processes, our approach eliminates the training-sampling gap, mitigates information leakage by treating the training as a direct mapping from pure Gaussian noise to the target data distribution, and supports the incorporation of perceptual and adversarial losses to enhance image fidelity and semantic consistency.
Our experiments on COCO30K and HW30K benchmarks demonstrate the effectiveness of the proposed method, achieving superior Fréchet Inception Distance (FID) and CLIP scores. These results highlight the robustness of end-to-end training paradigm, advancing diffusion-based generative models toward more practical and high-quality solutions.

{
    \small
    \bibliographystyle{ieee_fullname}
    \bibliography{ref}

\begin{thebibliography}{10}\itemsep=-1pt

\bibitem{bao2022analytic}
Fan Bao, Chongxuan Li, Jun Zhu, and Bo Zhang.
\newblock Analytic-dpm: an analytic estimate of the optimal reverse variance in diffusion probabilistic models.
\newblock {\em arXiv preprint arXiv:2201.06503}, 2022.

\bibitem{betker2023improving}
James Betker, Gabriel Goh, Li Jing, Tim Brooks, Jianfeng Wang, Linjie Li, Long Ouyang, Juntang Zhuang, Joyce Lee, Yufei Guo, et~al.
\newblock Improving image generation with better captions.
\newblock {\em Computer Science. https://cdn. openai. com/papers/dall-e-3. pdf}, 2(3):8, 2023.

\bibitem{carlini2024poisoning_coyo}
Nicholas Carlini, Matthew Jagielski, Christopher~A Choquette-Choo, Daniel Paleka, Will Pearce, Hyrum Anderson, Andreas Terzis, Kurt Thomas, and Florian Tram{\`e}r.
\newblock Poisoning web-scale training datasets is practical.
\newblock In {\em 2024 IEEE Symposium on Security and Privacy (SP)}, pages 407--425. IEEE, 2024.

\bibitem{chen2024pixart_delta}
Junsong Chen, Yue Wu, Simian Luo, Enze Xie, Sayak Paul, Ping Luo, Hang Zhao, and Zhenguo Li.
\newblock Pixart-$\{$$\backslash$delta$\}$: Fast and controllable image generation with latent consistency models.
\newblock {\em arXiv preprint arXiv:2401.05252}, 2024.

\bibitem{esser2403scaling}
Patrick Esser, Sumith Kulal, Andreas Blattmann, Rahim Entezari, Jonas M{\"u}ller, Harry Saini, Yam Levi, Dominik Lorenz, Axel Sauer, Frederic Boesel, et~al.
\newblock Scaling rectified flow transformers for high-resolution image synthesis, 2024.
\newblock {\em URL https://arxiv. org/abs/2403.03206}, 2.

\bibitem{goodfellow2020generative}
Ian Goodfellow, Jean Pouget-Abadie, Mehdi Mirza, Bing Xu, David Warde-Farley, Sherjil Ozair, Aaron Courville, and Yoshua Bengio.
\newblock Generative adversarial networks.
\newblock {\em Communications of the ACM}, 63(11):139--144, 2020.

\bibitem{gupta2024photorealistic}
Agrim Gupta, Lijun Yu, Kihyuk Sohn, Xiuye Gu, Meera Hahn, Fei-Fei Li, Irfan Essa, Lu Jiang, and Jos{\'e} Lezama.
\newblock Photorealistic video generation with diffusion models.
\newblock In {\em European Conference on Computer Vision}, pages 393--411. Springer, 2024.

\bibitem{he2024cameractrl}
Hao He, Yinghao Xu, Yuwei Guo, Gordon Wetzstein, Bo Dai, Hongsheng Li, and Ceyuan Yang.
\newblock Cameractrl: Enabling camera control for text-to-video generation.
\newblock {\em arXiv preprint arXiv:2404.02101}, 2024.

\bibitem{heusel2017gans_fid}
Martin Heusel, Hubert Ramsauer, Thomas Unterthiner, Bernhard Nessler, and Sepp Hochreiter.
\newblock Gans trained by a two time-scale update rule converge to a local nash equilibrium.
\newblock {\em Advances in neural information processing systems}, 30, 2017.

\bibitem{ho2020denoising}
Jonathan Ho, Ajay Jain, and Pieter Abbeel.
\newblock Denoising diffusion probabilistic models.
\newblock {\em Advances in neural information processing systems}, 33:6840--6851, 2020.

\bibitem{ho2022video}
Jonathan Ho, Tim Salimans, Alexey Gritsenko, William Chan, Mohammad Norouzi, and David~J Fleet.
\newblock Video diffusion models.
\newblock {\em Advances in Neural Information Processing Systems}, 35:8633--8646, 2022.

\bibitem{hoppe2022diffusion}
Tobias H{\"o}ppe, Arash Mehrjou, Stefan Bauer, Didrik Nielsen, and Andrea Dittadi.
\newblock Diffusion models for video prediction and infilling.
\newblock {\em arXiv preprint arXiv:2206.07696}, 2022.

\bibitem{hu2024animate}
Li Hu.
\newblock Animate anyone: Consistent and controllable image-to-video synthesis for character animation.
\newblock In {\em Proceedings of the IEEE/CVF Conference on Computer Vision and Pattern Recognition}, pages 8153--8163, 2024.

\bibitem{karras2022elucidating}
Tero Karras, Miika Aittala, Timo Aila, and Samuli Laine.
\newblock Elucidating the design space of diffusion-based generative models.
\newblock {\em Advances in neural information processing systems}, 35:26565--26577, 2022.

\bibitem{karras2024analyzing}
Tero Karras, Miika Aittala, Jaakko Lehtinen, Janne Hellsten, Timo Aila, and Samuli Laine.
\newblock Analyzing and improving the training dynamics of diffusion models.
\newblock In {\em Proceedings of the IEEE/CVF Conference on Computer Vision and Pattern Recognition}, pages 24174--24184, 2024.

\bibitem{kingma2013auto}
Diederik~P Kingma.
\newblock Auto-encoding variational bayes.
\newblock {\em arXiv preprint arXiv:1312.6114}, 2013.

\bibitem{ledig2017photo}
Christian Ledig, Lucas Theis, Ferenc Husz{\'a}r, Jose Caballero, Andrew Cunningham, Alejandro Acosta, Andrew Aitken, Alykhan Tejani, Johannes Totz, Zehan Wang, et~al.
\newblock Photo-realistic single image super-resolution using a generative adversarial network.
\newblock In {\em Proceedings of the IEEE conference on computer vision and pattern recognition}, pages 4681--4690, 2017.

\bibitem{lin2024sdxl}
Shanchuan Lin, Anran Wang, and Xiao Yang.
\newblock Sdxl-lightning: Progressive adversarial diffusion distillation.
\newblock {\em arXiv preprint arXiv:2402.13929}, 2024.

\bibitem{lin2014microsoft}
Tsung-Yi Lin, Michael Maire, Serge Belongie, James Hays, Pietro Perona, Deva Ramanan, Piotr Doll{\'a}r, and C~Lawrence Zitnick.
\newblock Microsoft coco: Common objects in context.
\newblock In {\em Computer Vision--ECCV 2014: 13th European Conference, Zurich, Switzerland, September 6-12, 2014, Proceedings, Part V 13}, pages 740--755. Springer, 2014.

\bibitem{liu2024visual}
Haotian Liu, Chunyuan Li, Qingyang Wu, and Yong~Jae Lee.
\newblock Visual instruction tuning.
\newblock {\em Advances in neural information processing systems}, 36, 2024.

\bibitem{liu2021generic}
Yifan Liu, Hao Chen, Yu Chen, Wei Yin, and Chunhua Shen.
\newblock Generic perceptual loss for modeling structured output dependencies.
\newblock In {\em Proceedings of the IEEE/CVF Conference on Computer Vision and Pattern Recognition}, pages 5424--5432, 2021.

\bibitem{lu2022dpm}
Cheng Lu, Yuhao Zhou, Fan Bao, Jianfei Chen, Chongxuan Li, and Jun Zhu.
\newblock Dpm-solver: A fast ode solver for diffusion probabilistic model sampling in around 10 steps.
\newblock {\em Advances in Neural Information Processing Systems}, 35:5775--5787, 2022.

\bibitem{luo2023latent}
Simian Luo, Yiqin Tan, Longbo Huang, Jian Li, and Hang Zhao.
\newblock Latent consistency models: Synthesizing high-resolution images with few-step inference.
\newblock {\em arXiv preprint arXiv:2310.04378}, 2023.

\bibitem{ma2024latte}
Xin Ma, Yaohui Wang, Gengyun Jia, Xinyuan Chen, Ziwei Liu, Yuan-Fang Li, Cunjian Chen, and Yu Qiao.
\newblock Latte: Latent diffusion transformer for video generation.
\newblock {\em arXiv preprint arXiv:2401.03048}, 2024.

\bibitem{meng2023distillation}
Chenlin Meng, Robin Rombach, Ruiqi Gao, Diederik Kingma, Stefano Ermon, Jonathan Ho, and Tim Salimans.
\newblock On distillation of guided diffusion models.
\newblock In {\em Proceedings of the IEEE/CVF Conference on Computer Vision and Pattern Recognition}, pages 14297--14306, 2023.

\bibitem{nichol2021improved}
Alexander~Quinn Nichol and Prafulla Dhariwal.
\newblock Improved denoising diffusion probabilistic models.
\newblock In {\em International conference on machine learning}, pages 8162--8171. PMLR, 2021.

\bibitem{peebles2023scalable}
William Peebles and Saining Xie.
\newblock Scalable diffusion models with transformers.
\newblock In {\em Proceedings of the IEEE/CVF international conference on computer vision}, pages 4195--4205, 2023.

\bibitem{sdxl}
Dustin Podell, Zion English, Kyle Lacey, Andreas Blattmann, Tim Dockhorn, Jonas M{\"u}ller, Joe Penna, and Robin Rombach.
\newblock Sdxl: Improving latent diffusion models for high-resolution image synthesis.
\newblock {\em arXiv preprint arXiv:2307.01952}, 2023.

\bibitem{poole2022dreamfusion}
Ben Poole, Ajay Jain, Jonathan~T Barron, and Ben Mildenhall.
\newblock Dreamfusion: Text-to-3d using 2d diffusion.
\newblock {\em arXiv preprint arXiv:2209.14988}, 2022.

\bibitem{radford2021clip}
Alec Radford, Jong~Wook Kim, Chris Hallacy, Aditya Ramesh, Gabriel Goh, Sandhini Agarwal, Girish Sastry, Amanda Askell, Pamela Mishkin, Jack Clark, et~al.
\newblock Learning transferable visual models from natural language supervision.
\newblock In {\em International conference on machine learning}, pages 8748--8763. PMLR, 2021.

\bibitem{ramesh2022hierarchical}
Aditya Ramesh, Prafulla Dhariwal, Alex Nichol, Casey Chu, and Mark Chen.
\newblock Hierarchical text-conditional image generation with clip latents.
\newblock {\em arXiv preprint arXiv:2204.06125}, 1(2):3, 2022.

\bibitem{rombach2022high}
Robin Rombach, Andreas Blattmann, Dominik Lorenz, Patrick Esser, and Bj{\"o}rn Ommer.
\newblock High-resolution image synthesis with latent diffusion models.
\newblock In {\em Proceedings of the IEEE/CVF conference on computer vision and pattern recognition}, pages 10684--10695, 2022.

\bibitem{saharia2022photorealistic}
Chitwan Saharia, William Chan, Saurabh Saxena, Lala Li, Jay Whang, Emily~L Denton, Kamyar Ghasemipour, Raphael Gontijo~Lopes, Burcu Karagol~Ayan, Tim Salimans, et~al.
\newblock Photorealistic text-to-image diffusion models with deep language understanding.
\newblock {\em Advances in neural information processing systems}, 35:36479--36494, 2022.

\bibitem{salimans2022progressive}
Tim Salimans and Jonathan Ho.
\newblock Progressive distillation for fast sampling of diffusion models.
\newblock {\em arXiv preprint arXiv:2202.00512}, 2022.

\bibitem{singer2022make}
Uriel Singer, Adam Polyak, Thomas Hayes, Xi Yin, Jie An, Songyang Zhang, Qiyuan Hu, Harry Yang, Oron Ashual, Oran Gafni, et~al.
\newblock Make-a-video: Text-to-video generation without text-video data.
\newblock {\em arXiv preprint arXiv:2209.14792}, 2022.

\bibitem{songdenoising}
Jiaming Song, Chenlin Meng, and Stefano Ermon.
\newblock Denoising diffusion implicit models.
\newblock In {\em International Conference on Learning Representations}, 2021.

\bibitem{song2023improved}
Yang Song and Prafulla Dhariwal.
\newblock Improved techniques for training consistency models.
\newblock {\em arXiv preprint arXiv:2310.14189}, 2023.

\bibitem{song2023consistency}
Yang Song, Prafulla Dhariwal, Mark Chen, and Ilya Sutskever.
\newblock Consistency models.
\newblock {\em arXiv preprint arXiv:2303.01469}, 2023.

\bibitem{song2020score}
Yang Song, Jascha Sohl-Dickstein, Diederik~P Kingma, Abhishek Kumar, Stefano Ermon, and Ben Poole.
\newblock Score-based generative modeling through stochastic differential equations.
\newblock {\em arXiv preprint arXiv:2011.13456}, 2020.

\bibitem{sun2024hunyuan}
Xingwu Sun, Yanfeng Chen, Yiqing Huang, Ruobing Xie, Jiaqi Zhu, Kai Zhang, Shuaipeng Li, Zhen Yang, Jonny Han, Xiaobo Shu, et~al.
\newblock Hunyuan-large: An open-source moe model with 52 billion activated parameters by tencent.
\newblock {\em arXiv preprint arXiv:2411.02265}, 2024.

\bibitem{tan2024empirical}
Zhiyu Tan, Mengping Yang, Luozheng Qin, Hao Yang, Ye Qian, Qiang Zhou, Cheng Zhang, and Hao Li.
\newblock An empirical study and analysis of text-to-image generation using large language model-powered textual representation.
\newblock In {\em European Conference on Computer Vision}, pages 472--489. Springer, 2024.

\bibitem{wang2018esrgan}
Xintao Wang, Ke Yu, Shixiang Wu, Jinjin Gu, Yihao Liu, Chao Dong, Yu Qiao, and Chen Change~Loy.
\newblock Esrgan: Enhanced super-resolution generative adversarial networks.
\newblock In {\em Proceedings of the European conference on computer vision (ECCV) workshops}, pages 0--0, 2018.

\bibitem{xiao2021tackling}
Zhisheng Xiao, Karsten Kreis, and Arash Vahdat.
\newblock Tackling the generative learning trilemma with denoising diffusion gans.
\newblock {\em arXiv preprint arXiv:2112.07804}, 2021.

\bibitem{yang2024cogvideox}
Zhuoyi Yang, Jiayan Teng, Wendi Zheng, Ming Ding, Shiyu Huang, Jiazheng Xu, Yuanming Yang, Wenyi Hong, Xiaohan Zhang, Guanyu Feng, et~al.
\newblock Cogvideox: Text-to-video diffusion models with an expert transformer.
\newblock {\em arXiv preprint arXiv:2408.06072}, 2024.

\bibitem{zheng2024cami2v}
Guangcong Zheng, Teng Li, Rui Jiang, Yehao Lu, Tao Wu, and Xi Li.
\newblock Cami2v: Camera-controlled image-to-video diffusion model.
\newblock {\em arXiv preprint arXiv:2410.15957}, 2024.

\bibitem{zhou2024fast}
Zhenyu Zhou, Defang Chen, Can Wang, and Chun Chen.
\newblock Fast ode-based sampling for diffusion models in around 5 steps.
\newblock In {\em Proceedings of the IEEE/CVF Conference on Computer Vision and Pattern Recognition}, pages 7777--7786, 2024.

\end{thebibliography}
}

\newpage 
\appendix

\section{Limitations and Future Works}

\subsection{Limitations.} Despite demonstrating substantial improvements in generative performance, our proposed method has several limitations. Firstly, our current end-to-end training strategy primarily involves fine-tuning from pretrained diffusion models. We have not yet explored training entirely from random initialization, which could potentially lead to differences in convergence behaviors and ultimate generative performance. Secondly, the proposed method's computational complexity remains relatively high, especially in scenarios involving large-scale datasets or high-resolution image synthesis, potentially limiting its scalability and practical deployment. Lastly, while our experiments provide comprehensive validation on benchmark datasets such as COCO30K and HW30K, the generalization capability to significantly larger and more diverse datasets like ImageNet has not been fully investigated.

\subsection{Future Work.} Addressing these limitations opens several promising research directions. Future research should investigate the feasibility and effectiveness of end-to-end diffusion training from scratch (random initialization), systematically evaluating its impact on model convergence, robustness, and final image quality. Another valuable direction would be improving computational efficiency through architectural innovations, advanced sampling acceleration methods, and techniques such as knowledge distillation, thereby making our model more suitable for resource-constrained environments. Moreover, extending our method to more complex and diverse generative tasks, including high-resolution image generation, temporally consistent video synthesis, and multi-modal generation scenarios, would further validate and expand the method’s applicability and robustness. Lastly, comprehensive analysis on larger-scale datasets, such as ImageNet or domain-specific datasets, could significantly enhance the understanding and validate the generalizability of our end-to-end learning framework, paving the way toward broader adoption and practical utility of diffusion-based generative models.

\section{Ethical and Social Impacts}
\subsection{Ethical Considerations}

\mypara{Content Authenticity and Potential Misuse.} Generative models, such as the diffusion framework presented in this study, have significantly enhanced the realism and quality of synthesized visual content. However, these advancements pose potential ethical risks related to content authenticity. There exists a heightened risk that generated images could be misused for creating deceptive content, including misinformation or deepfake imagery, potentially undermining trust in digital media.

\mypara{Privacy Concerns.} High-quality generative methods could inadvertently generate realistic depictions of individuals without their explicit consent. Such misuse can lead to serious privacy infringements, identity theft, or unauthorized impersonation, highlighting a critical area requiring vigilant ethical oversight and robust safeguards.

\mypara{Bias and Fairness.} Generative models may inherit and amplify biases present within training datasets, resulting in biased or unfair representations across demographic groups. This amplification could reinforce harmful stereotypes or unfairly disadvantage certain populations, thereby necessitating rigorous assessment and proactive strategies to mitigate biases.

\subsection{Social Implications}

\mypara{Influence on Creative Professions.} Our approach, characterized by improved realism and efficiency, may profoundly influence industries reliant on creative and visual content creation, such as digital arts, photography, design, and entertainment. While enhancing productivity and creativity, these technological advancements may also disrupt traditional job markets, necessitating adaptive measures and new professional training paradigms.

\mypara{Democratization and Accessibility.} By significantly lowering the barriers to entry for creating sophisticated visual content, our method contributes to the democratization of generative technology. Wider accessibility empowers individuals without extensive technical expertise, promoting innovation and fostering greater inclusivity across diverse user groups.

\mypara{Educational and Cultural Opportunities.} Generative methods like ours present opportunities for educational innovation and cultural preservation. For instance, they can facilitate engaging learning experiences, aid in the restoration and digitization of cultural heritage, and support collaborative creative endeavors, thereby enriching societal and cultural dynamics.

\subsection{Mitigation and Recommendations}

\mypara{Development of Responsible Use Guidelines.} We advocate establishing comprehensive ethical guidelines for the responsible deployment and use of generative models. Clearly articulated standards can help developers and users navigate ethical dilemmas and minimize misuse.

\mypara{Transparency in Generated Content.} Promoting transparency through mandatory labeling or digital watermarking of generated content can help distinguish authentic from synthetic media. Such practices enhance trustworthiness and allow users to critically evaluate digital content.

\mypara{Strategies for Bias Mitigation.} Future research should prioritize developing robust methodologies to detect and mitigate biases within generative models systematically. Continuous monitoring, diverse training datasets, and fairness-aware learning algorithms represent effective approaches toward achieving equitable and unbiased generative outcomes.

\section{More Implementation Details}

\subsection{Sampling Details}
In the sampling phase of our diffusion model, we utilized the LCMScheduler~\cite{luo2023latent}, aligned with the approach introduced by PixArt-\(\delta\)~\cite{chen2024pixart_delta}. Specifically, we employed a linear noise schedule, initiating with a noise level \( \beta_{0}=0.0001 \) and concluding with \( \beta_{T}=0.02 \). This linear interpolation approach established a continuous sequence of noise intensities over \( T=1000 \) training steps. Due to the intrinsic design of the LCM architecture, a guidance scale of 4.5 was inherently integrated into the base model. Therefore, during sampling, we explicitly set the classifier-free guidance (CFG) to zero to preserve consistency and effectively utilize the pretrained model's optimized performance.

\subsection{ Adversarial Training Details}
In our adversarial training framework, we designed a discriminator comprising four convolutional layers, each employing a $4\times4$ convolutional kernel. The initial three layers progressively downsample the spatial dimensions by a factor of two. Throughout the training process, the discriminator evaluates two distinct image types: synthesized images produced by the diffusion model, classified as negative samples, and authentic images drawn from the real dataset, classified as positive samples. The adversarial loss is integrated into the overall objective with a weight of 0.01.
Considering the discriminator's random initialization, we adopted an asymmetric training regimen. Specifically, we updated the discriminator five times per generator update. This strategy ensures more stable adversarial training dynamics and mitigates potential convergence issues stemming from an initially underperforming discriminator.


\section{More Analysis Results}


\subsection{Parameter Sensitivity}
\begin{figure}[htpb]
    \centering
    \includegraphics[width=\linewidth]{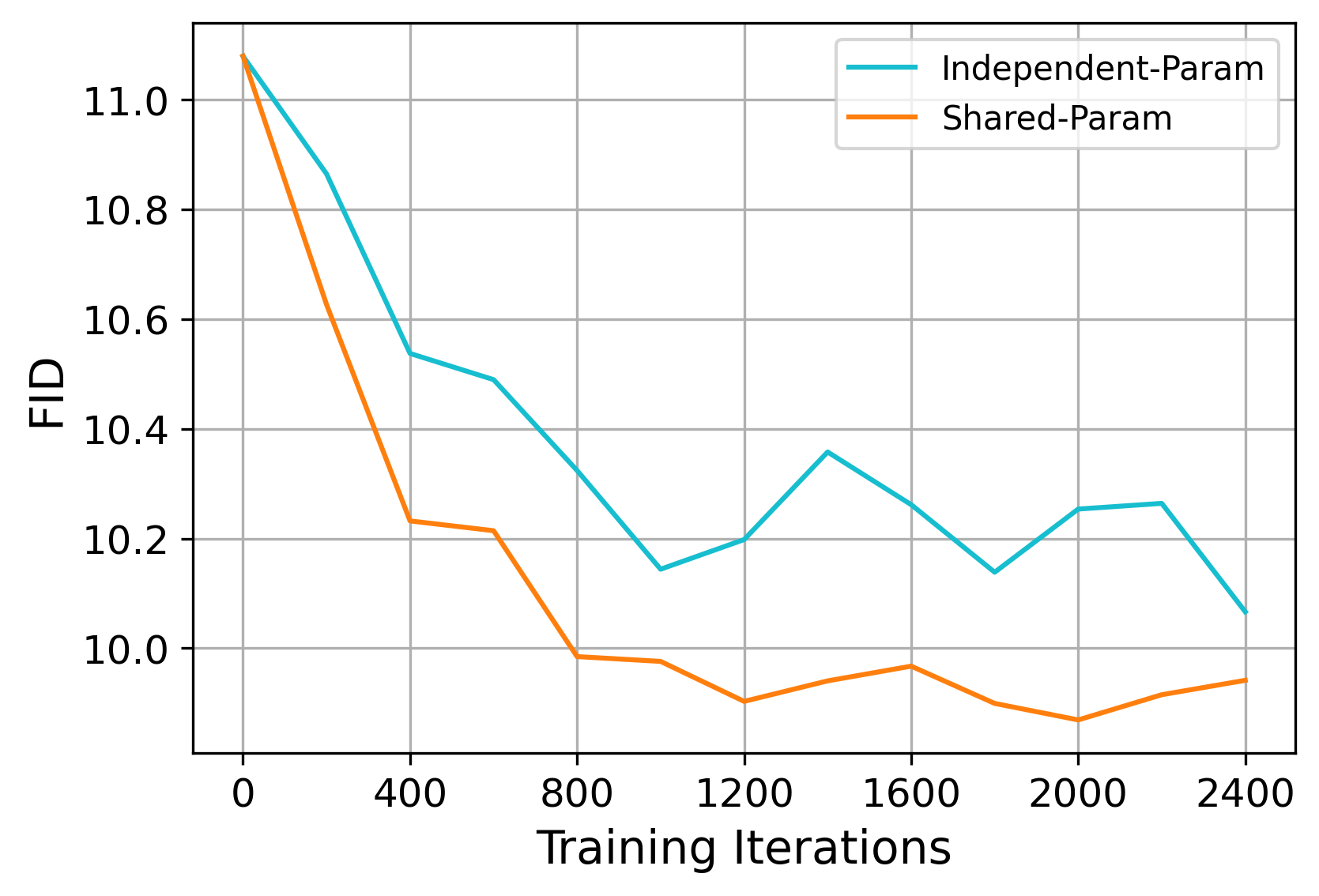}
    \caption{Comparison of training loss curves between parameter-sharing and parameter-independent configurations. The results demonstrate that parameter sharing consistently achieves lower loss values, highlighting its effectiveness in stabilizing training and improving overall model performance.}
    \label{fig:loss_param}
\end{figure}

To evaluate parameter sensitivity, traditional diffusion models utilize the same model for noise prediction across all time steps, meaning all noise prediction operations share identical parameters. To further investigate how parameter sharing influences model performance, we conducted comparative experiments between parameter-sharing and parameter-independent configurations. During training, the model was optimized using a combined loss comprising L2 and LPIPS metrics. Experimental results (shown in the figure\ref{fig:loss_param}) demonstrate that the parameter-sharing configuration notably outperforms the parameter-independent variant, highlighting the effectiveness of parameter sharing in enhancing overall model performance.

\section{Human Visual Evaluation}

\mypara{Data Preparation}
The human evaluation was conducted pairwise, with each participant systematically comparing 300 rigorously matched image pairs. To ensure a robust and representative evaluation dataset, we carefully selected 300 diverse prompts from the COCO dataset~\cite{lin2014microsoft}. PixArt-\(\delta\)~\cite{chen2024pixart_delta} served as our baseline model, while our proposed model was developed by fine-tuning PixArt-\(\delta\) using our novel training approach. Both models employed an identical sampling scheduler to maintain consistency and fairness in the comparison. Consequently, we obtained 300 carefully curated image pairs, each consisting of one image generated by our proposed model and the corresponding image produced by the baseline.

\begin{figure}[htpb]
    \centering
    \includegraphics[width=\linewidth]{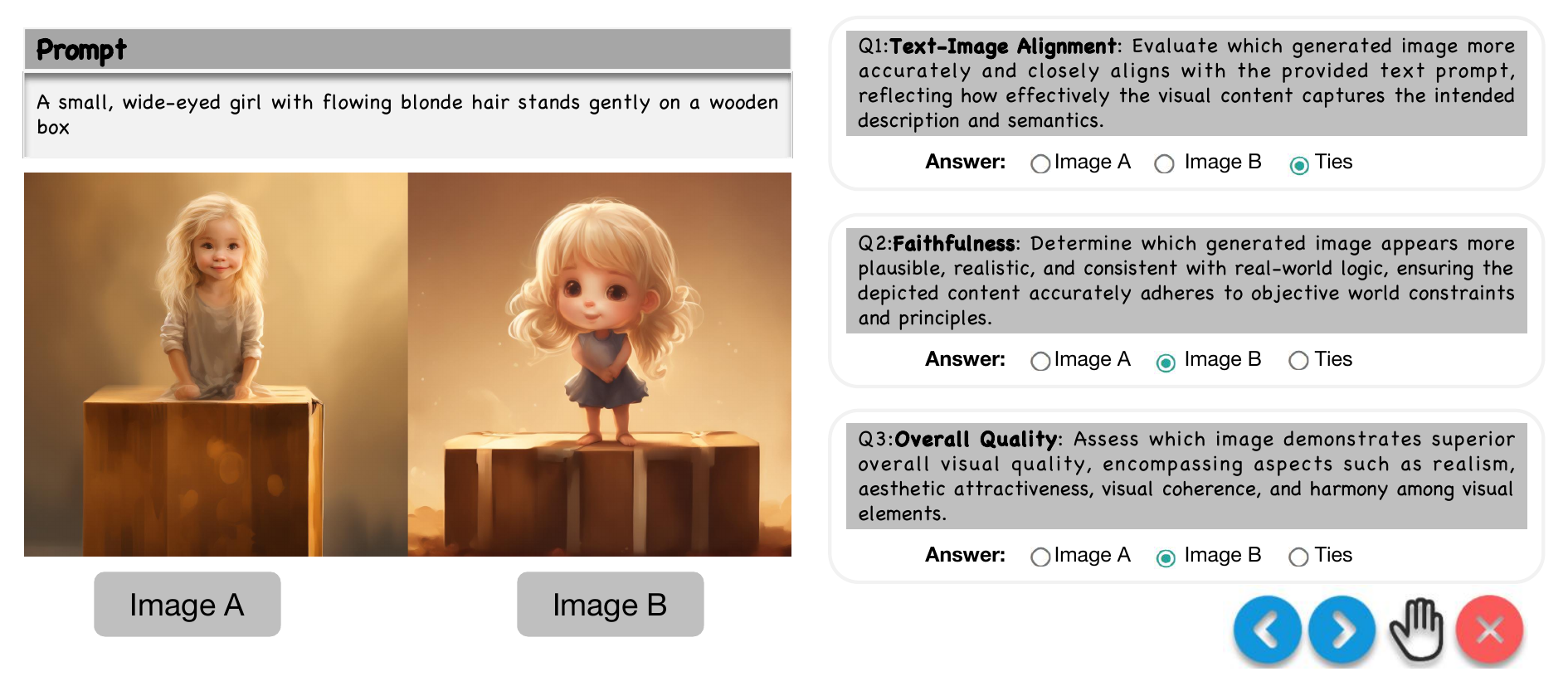}
    \caption{\textbf{User Interface Demonstration}: Our custom-designed user interface sequentially presents evaluators with pairs of images alongside the corresponding generation prompt. Additionally, we formulated three specific evaluation questions to comprehensively measure user preferences across three distinct dimensions.}
    \label{fig:interface}
\end{figure}

\mypara{User Interface and Evaluation Procedure}
To optimize evaluation efficiency and minimize potential biases, we designed an intuitive and user-friendly interface, illustrated in Figure~\ref{fig:interface}. Participants evaluated each image pair individually across three distinct dimensions: text-image alignment, faithfulness, and overall quality, indicating their preferences separately for each criterion. Additionally, user-friendly navigation via prominently positioned blue left and right arrow buttons facilitated seamless revisitation of previously assessed pairs. To ensure continuous participant support, a "raise hand" feature enabled immediate organizer assistance if evaluation-related questions arose.  Image pairs were randomized in presentation order, maintaining participant blindness to model identities throughout the evaluation.

\mypara{User Training and Guidelines}
Substantial preparatory measures were undertaken to ensure the fairness, consistency, and validity of our human evaluation. We initially implemented rigorous participant selection criteria, explicitly confirming their willingness to engage with potentially challenging visual content generated by evaluated models. From an initial cohort of 33 volunteers, we selected a diverse group of 10 participants comprising 3 graduate researchers specializing in text-to-image generation, 3 professional artists with demonstrated aesthetic sensitivity, and 4 non-expert participants. To guarantee evaluation accuracy and consistency, each selected participant underwent comprehensive training based on detailed guidelines meticulously prepared and presented in Table~\ref{tab:user_guidelines}. These guidelines outlined precise instructions for interface interaction and explicit evaluation criteria. Each participant received a Chinese-language version of the guidelines and attended dedicated training sessions emphasizing the evaluation's objectives, standards, and addressing individual questions comprehensively. This thorough training, coupled with a robust evaluation protocol and user-centric interface, ensured an effective, accurate, and bias-minimized evaluation process.

\begin{table}[t]
\footnotesize
\centering
\setlength\extrarowheight{2pt}
\caption{\textbf{User Guidelines for Human Evaluation}}
\begin{tabular}{|p{1\columnwidth}|}
\hline
\textbf{User Guidelines} \\
\hline
\textbf{Part I: User Interface Instructions} \\

1. Throughout the evaluation process, you will systematically compare 300 pairs of images using the provided user interface.

2. For each comparison, the interface will display:
\begin{itemize}
    \item A sequential number at the top-left corner to indicate your current progress.
    \item A pair of images generated by different models.
    \item A text prompt used for generating the images.
    \item Six buttons with various functions to facilitate your evaluation.
\end{itemize}

3. To indicate your preferred image, click the \textit{like} button located directly below the image.

4. If you have no clear preference between the two images, click the central red ``X'' button below the images.

5. After selecting your preference or indicating no preference, the interface will automatically move to the next image pair.

6. Navigation arrows located at the bottom-right corner of the interface allow you to revisit previously viewed pairs and revise your choices if necessary.

7. If you encounter any uncertainties or difficulties during the evaluation, please click the \textit{raise hand} button at the bottom-right corner, and assistance will be promptly provided.

\\ \hline
\textbf{Part II: Evaluation Criteria and Guidelines} \\

1. Generally, your evaluation should reflect your personal preference.

2. If you find it challenging to decide based solely on personal preference, we suggest using the following criteria:

\textbf{Text-Image Alignment}: Determine whether the generated image accurately aligns with the provided reference prompt.

\textbf{Faithfulness}: Assess if the generated image appears plausible and realistically conforms to the laws of the real world.

\textbf{Overall Quality}: Consider the general visual quality of the image, including its realism, aesthetic appeal, and coherence.

3. Initially, carefully evaluate 30 image pairs to establish a consistent evaluation standard. Afterwards, proceed to assess all remaining image pairs sequentially.

4. Should you feel uncertain or confused at any stage, use the \textit{raise hand} button at the bottom-right corner for assistance.

5. Upon completing the evaluation of all 300 image pairs, submit your results by clicking the right arrow button.

6. After submitting your evaluation, your task is complete. We sincerely appreciate your contribution to this research project.

\\ \hline
\end{tabular}
\label{tab:user_guidelines}
\end{table}

\mypara{Evaluation Results}

\begin{figure}[htbp]
    \centering
    \includegraphics[width=\linewidth]{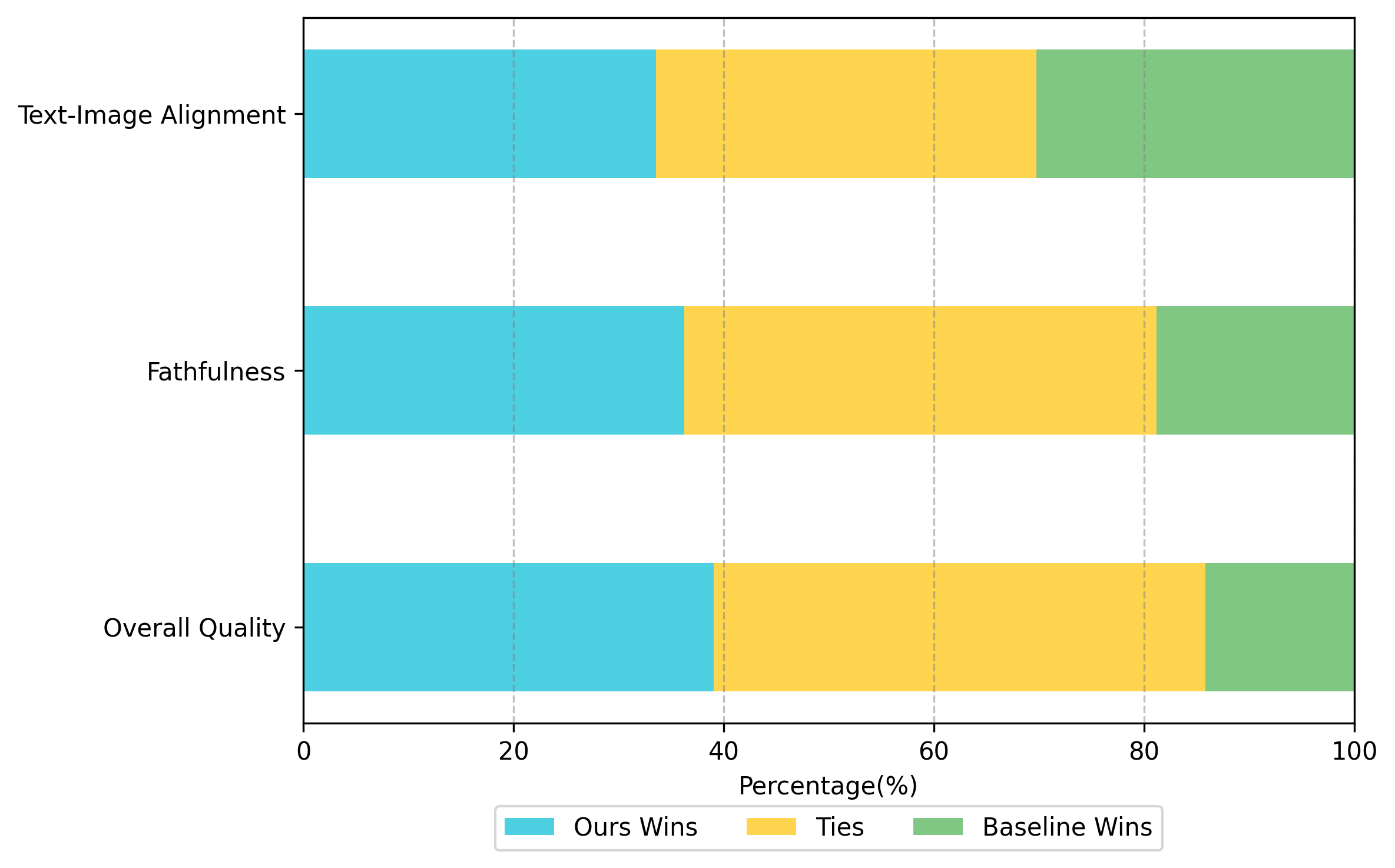}
    \caption{\textbf{Human Evaluation Results.} Our model consistently receives higher preference ratings from human evaluators, demonstrating its superior capability in generating images with enhanced visual quality.}
    \label{fig:human_eval}
\end{figure}

To thoroughly understand human preferences regarding \ourmethod, we conducted an extensive human evaluation comparing our model against PixArt-\(\delta\)~\cite{chen2024pixart_delta} through a pairwise study. Specifically, participants assessed the generated images across three critical dimensions: text-image alignment, faithfulness, and overall quality. Detailed definitions for these evaluation criteria can be found in Table~\ref{tab:user_guidelines}. As clearly illustrated in Figure~\ref{fig:human_eval}, the evaluation results consistently demonstrate that our proposed \ourmethod significantly outperforms the PixArt-\(\delta\) baseline in all three dimensions. This compelling evidence underscores the effectiveness of our approach in generating images that align closely with human preferences and exhibit superior visual fidelity and coherence.

\section{More Qualitative Results}
\begin{figure*}[htbp]
  \centering
  \includegraphics[width=\textwidth]{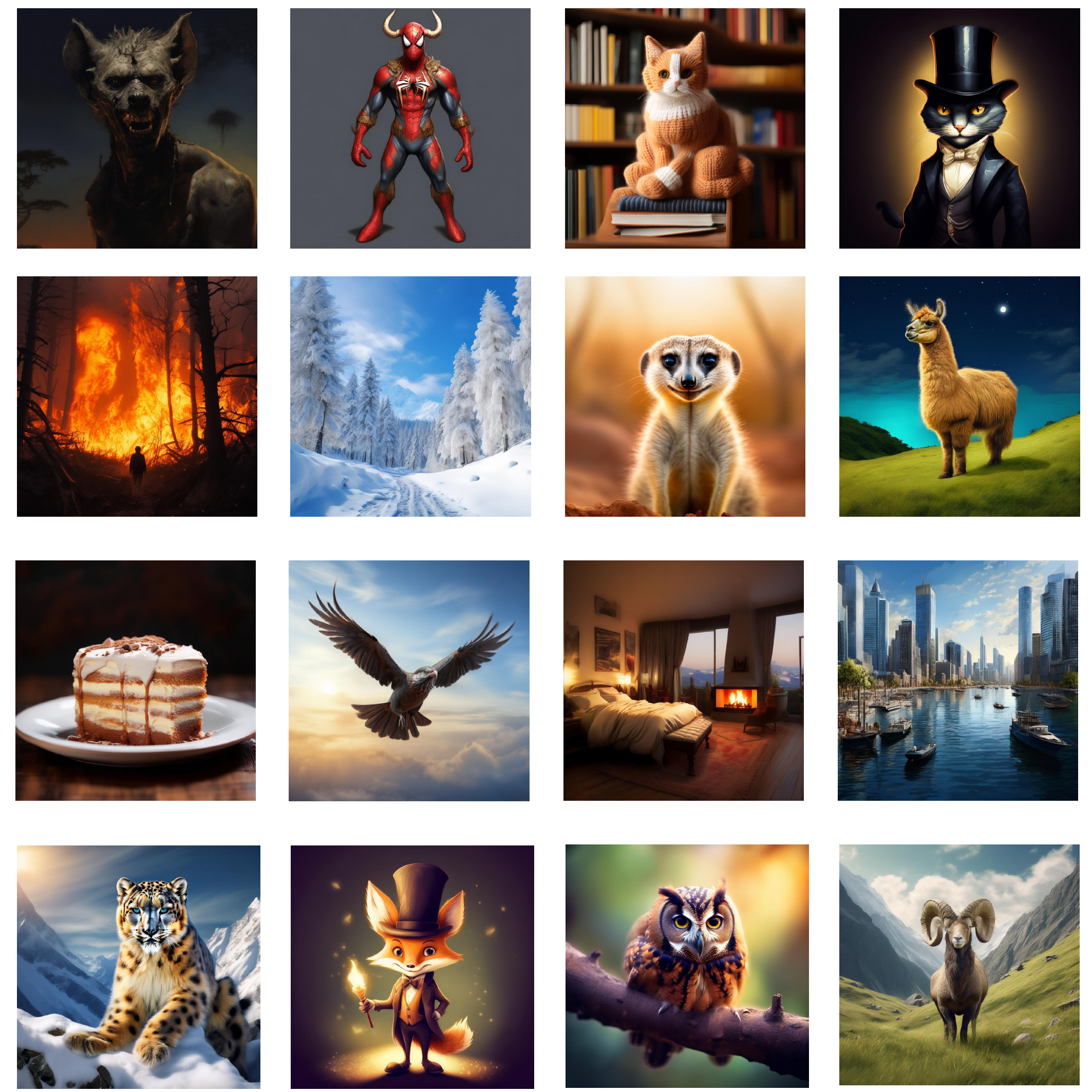}
  \caption{Additional qualitative results demonstrating the effectiveness of our text-to-image generation method. The images showcase high-fidelity details, strong semantic alignment with the input prompts, and a diverse range of visual styles.}
  \label{fig:qualitative_results_supp}
\end{figure*}

To further illustrate the effectiveness of our text-to-image generation method, we provide additional qualitative results in Figure~\ref{fig:qualitative_results_supp}. These examples highlight the model's ability to generate high-fidelity images with fine details, strong semantic alignment with input prompts, and diverse visual styles. The results demonstrate the robustness and versatility of our approach across different subjects and artistic styles, reinforcing its effectiveness in generating high-quality, text-aligned images.

\end{document}